\documentclass{article}

\usepackage{template}

\usepackage[utf8]{inputenc} % allow utf-8 input
\usepackage[T1]{fontenc}    % use 8-bit T1 fonts
\usepackage{hyperref}       % hyperlinks
\usepackage{url}            % simple URL typesetting
\usepackage{booktabs}       % professional-quality tables
\usepackage{amsfonts}       % blackboard math symbols
\usepackage{nicefrac}       % compact symbols for 1/2, etc.
\usepackage{microtype}      % microtypography
\usepackage{lipsum}		% Can be removed after putting your text content

\usepackage{times}
\usepackage{latexsym}
\usepackage{graphicx}
\usepackage{tabulary}

\title{Medical Segment Coloring of Clinical Notes}

\author{
  Maha Alkhairy \\
  Northeastern University, Computer Science\\
   {\tt alkhairy.m@northeastern.edu} \\
}

\date{}

\renewcommand{\shorttitle}{Medical Segment Coloring of Clinical Notes}

\begin{document}
\maketitle

\begin{abstract}

This paper proposes a deep learning-based method to identify the segments of a clinical note corresponding to ICD-9 broad categories which are further color-coded with respect to 17 ICD-9 categories. 

The proposed Medical Segment Colorer (MSC) architecture  is a pipeline framework that works in three stages: (1) word categorization, (2) phrase allocation, and (3) document classification. MSC uses gated recurrent unit neural networks (GRUs) to map from an input document to word multi-labels to phrase allocations, and uses statistical median to map phrase allocation to document multi-label. We compute variable length segment coloring from overlapping phrase allocation probabilities. These cross-level bidirectional contextual links identify adaptive context and then produce segment coloring. 

We train and evaluate MSC using the document labeled MIMIC-III clinical notes. Training is conducted solely using document multi-labels without any information on phrases, segments, or words. In addition to coloring a clinical note, MSC generates as byproducts document multi-labeling and word tagging -- creation of ICD9 category keyword lists based on segment coloring.

Performance comparison of  MSC byproduct document multi-labels versus methods whose purpose is to produce justifiable document multi-labels is $64\%$ vs $52.4\%$ micro-average F1-score against the CAML (CNN attention multi label) method.

For evaluation of MSC segment coloring results, medical practitioners independently assigned the colors to broad ICD9 categories given a sample of 40 colored notes and a sample of 50 words related to each category based on the word tags. Binary scoring of this evaluation has a median value of $83.3\%$ and mean of $63.7\%$.

\end{abstract}

% keywords can be removed

% \twocolumn
\section{Introduction}

\begin{figure*}[!hbt]
  
  \centering
%  trim={<left> <lower> <right> <upper>}
    \includegraphics[trim={0 0 0 3.1cm},clip, width=1.0\linewidth]{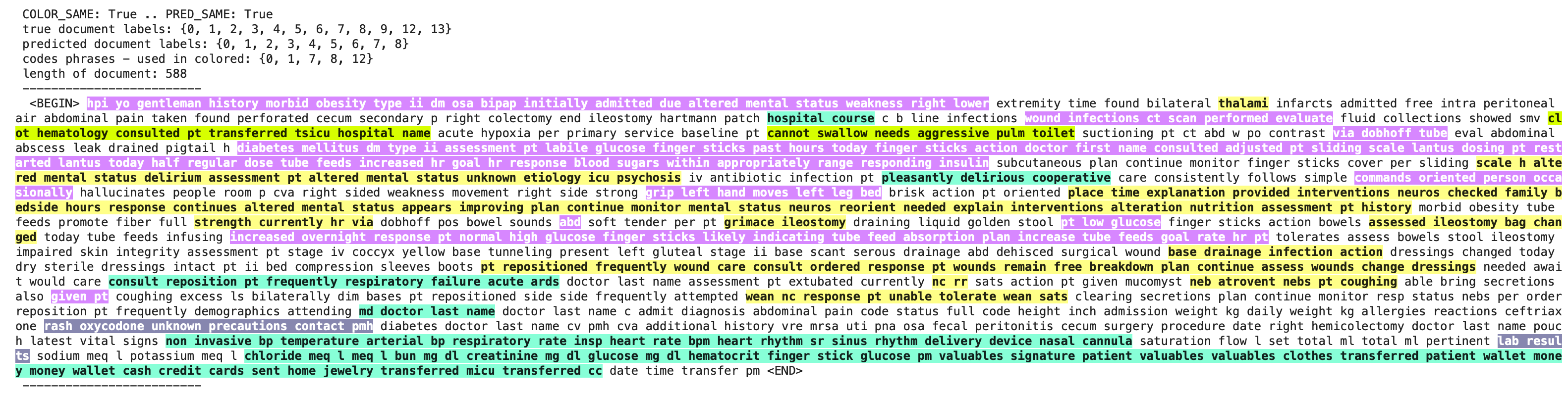}
    
    \includegraphics[trim={0 0 0 3.5cm},clip,width=1.0\linewidth]{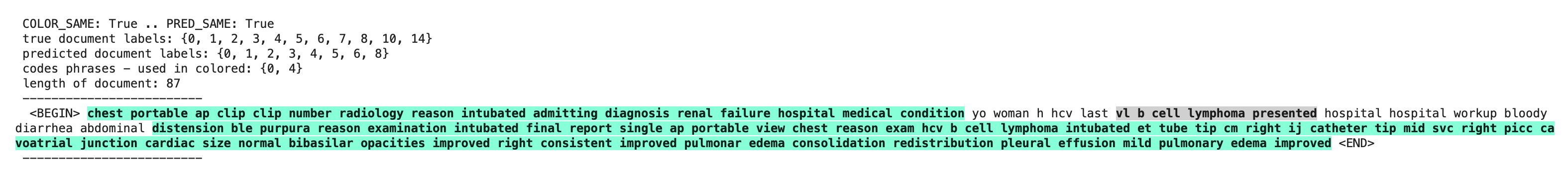}
    
    \includegraphics[trim={0 0 0 3.1cm},clip,width=1.0\linewidth]{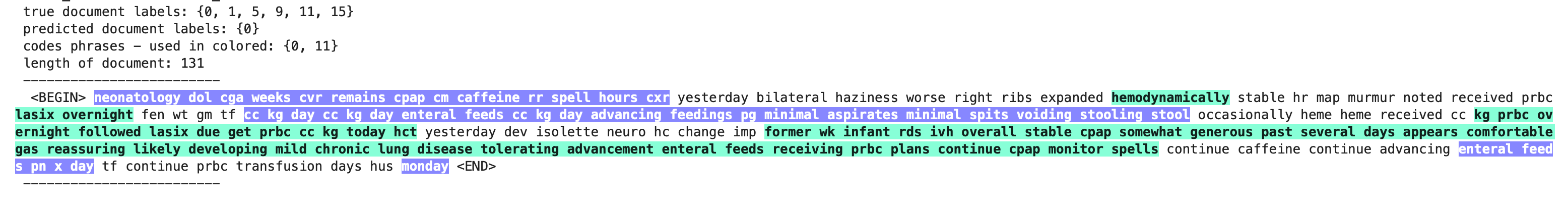}
    
  \caption{Three clinical notes with colored segments corresponding to ICD9 broad categories -- sample outputs of MSC run on preprocessed test clinical notes.}
  \label{fig:example_colored}
\end{figure*}

\begin{figure*}[!hbt]
  \centering
    \includegraphics[width=\linewidth]{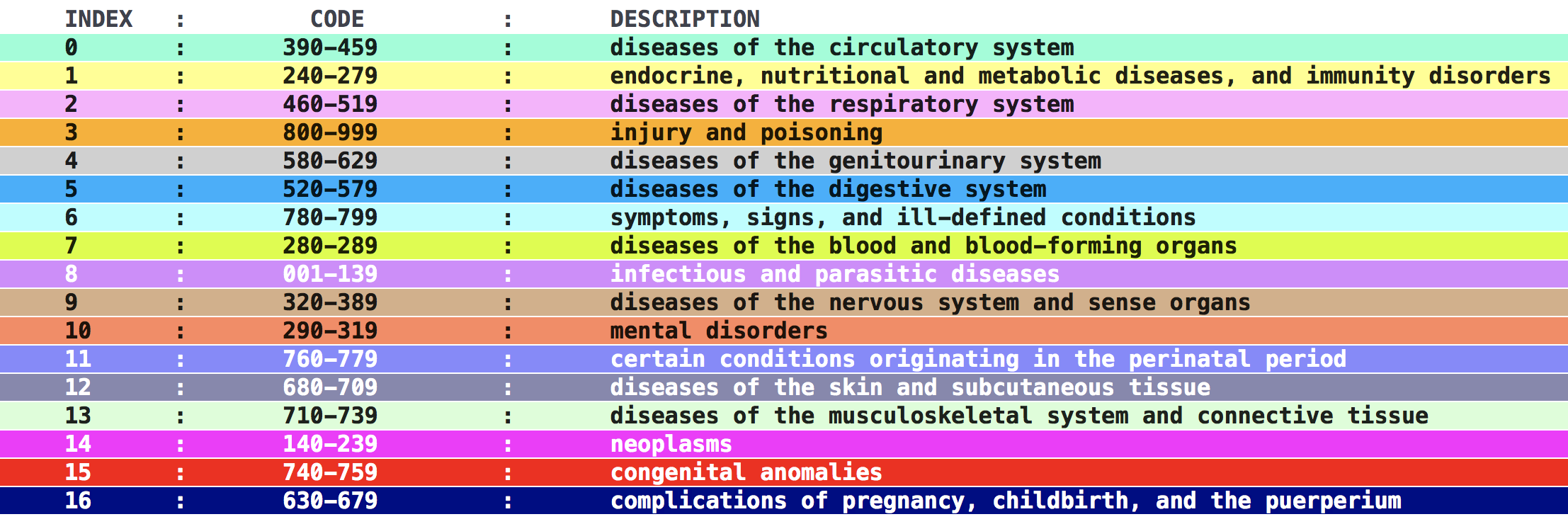}
  \caption{Legend: index and color are keys, and ICD9 category codes with descriptions are values. Index and color assignment to ICD9 categories is arbitrary. The values for the keys are not known to the MSC model and the medical practitioners used as evaluators.}
  \label{fig:legend}
\end{figure*}

A variety of clinical notes are produced during the stay and discharge of patients. A note may be a discharge summary, nursing note, or physician report and range in length from  a few phrases to thousands of words. Some clinical notes are multi-labeled with unified ICD9 codes \cite{slee1978international} for purposes of financial billing and insurance. These multi-level codes classify disease and related health problems with 17 codes at the top level categorization into broad diseases and specialties. Although clinical notes provide a wealth of information to physicians and analysts, the content may relate to differing specializations and key information may be overlooked, specially in cases of long histories for a patient or time constrained actions.

However, a clinical note that has medical segments relevant to each specialization colored with a distinct color would create focus on specific texts of concern to a specialist within the context of the entire clinical note. The clinical note may be of arbitrary length or type such as nurse's note, radiology report, or discharge summary; and may contain multiple category colors. This is the motivation behind the problem that is formulated and addressed in this paper.

We design and build an end-to-end system that identifies, delimits, and colors clinical notes segments according to the broad ICD9 diagnosis categories they belong to -- leaving out generic segments. To our knowledge, this problem of medical segment coloring of clinical notes according to specialization has not been solved before. 

Our system architecture consists of serially linked word, phrase, and document modules that impose two-way constraints on one another thereby emphasizing context among other constraints, and pipelines an input document through these modules into word multi-labels, phrase allocations, and document multi-labels. Each module is built on the gated recurrent unit (GRU) \cite{cho2014learning} structure that imposes preceding and succeeding contexts. We  compute  variable  length  segment coloring from  overlapping phrase  allocation  probabilities.

The input of our system -- Medical Segment Colorer (MSC) -- is clinical note of any length and the output is colored segment without any padding or truncation of the document. The two byproducts of the model are document multi-labeling and word tagging, which is the creation of ICD9 category keyword lists based on segment coloring.  

MSC is trained on MIMIC-III multi-labeled documents indicating relevant broad ICD-9 diagnosis categories only without any descriptions. The input to training is the clinical note and its multi-labels at the document level. These labels are ICD-9 codes without any words, phrases or segments. Thus training for the segment coloring problem is conducted through the indirect relationship between documents and phrases and hence segments. This is an instance of transfer learning  \cite{pan2010survey} in which the model learns segment coloring from document multi-labels. 

To evaluate segment coloring and word tagging, medical practitioners were asked to determine the ICD9 category that each color in a colored clinical note corresponded to -- without being given the legend. A mean score of $63.7\%$ and median score of $83.3\%$ is achieved through binary matching of the practitioner's labels and the legend shown in Figure \ref{fig:legend}. Medical practitioners who evaluated noted the need, as many diagnoses are missed when dealing with long medical history for the patient.

\section{Problem Statement}
\label{problem_statement}

\begin{table*}[!hbt]
    \centering
    \begin{tabular}{ | l | l | l | l | l | l | }
    \hline
     Type & Count &  Type & Count &  Type & Count \\
    \hline
    Rehab Services & 5,431 & Pharmacy & 103  & ECG & 209,051  \\
    \hline
    Respiratory  & 31,739 & Consult & 98 &  Social Work & 2,670 \\
    \hline
    Nursing & 223,556 & Case Management  & 967 & Nursing/other & 822,497\\
    \hline
     Echo & 45,794 &  Radiology & 522,279 & General & 8,301\\
    \hline
    Discharge summary & 59,652 & Nutrition & 9418 & Physician  & 141,624\\
 
    \hline
    \end{tabular}           
    \caption{Statistics of MIMIC-III database note categories.}
    \label{tab:categories}
\end{table*}

The problem statement is: Given a clinical note of arbitrary length and of any type as input, identify and produce delimited colored segments in which a color corresponds to one of the 17 ICD-9 categories, without coloring non-medical segments. Figure \ref{fig:example_colored} illustrates a sample of clinical notes with colored medical segments. 

The granularity of coloring is at the top level of ICD-9 (e.g. 001-139, 240-279, ...) to match with specializations such as pulmonology and gastroenterology rather than at a lower level that is better suited for the different problem of multi-labelling documents for use in billing and insurance. The coloring problem is at a segment level whereas multi-labeling is at a document level.

The restriction on solving the problem is that no information on words, phrases, or segments are available for model estimation. The only inputs to training the model are clinical notes and their document level ICD9 codes. 

\section{Relevant Work}

We are unaware of work on the segment coloring problem. However, several works on multi-labeling clinical notes, specifically discharge summaries, using fine grained ICD9 codes (e.g 11.3) have been conducted. These works use a wide variety of methods and some use attention as a means to justify the classification. 
Multi-labelling methods used include rule-based \cite{crammer2007automatic} or rule based with boosting models \cite{goldstein2007three}, Naive Bayes \cite{Pakhomov2006ResearchPA}, SVM and Bayesian Ridge Regression  \cite{lita-etal-2008-large}, SVM using a variety of features \cite{patrick2007developing},   long-term short-term memory neural networks (LSTM) \cite{ayyar2016tagging}, and convolutional neural networks with attention (CAML) \cite{mullenbach-etal-2018-explainable,rios2018few}. The micro F1-score -- for multi-labeling note with fine-grained labels on the test set of MIMIC-III discharge summaries -- as demonstrated in the related papers for the deep learning approaches range from $52.4\%$ (CAML - diagnosis) to $70.8\%$ (LSTM).

In document multi-labeling, MSC achieves a $64\%$ micro F1-score on  MIMIC-III clinical notes of all types -- nursing notes, radiology reports, and discharge summaries. This compares favorably to CAML ($52.4\%$ micro F1-score) \cite{mullenbach-etal-2018-explainable} that is used to produce document multi-labels along with attention as an explanation for discharge summaries. 

\section{Data and Preprocessing}

We use ICD9 classification tree and the MIMIC-III database \cite{johnson2016mimic} of clinical notes that are annotated at the document level with multiple ICD9 fine grain labels. 

\begin{table}[!hbt]
\centering
{
\begin{tabular}{ | c | c | l | }
\hline
Index & Code & MSC Tagged Words\\
\hline
0 & 390-459 & subdural,\ arteries,\ hypertension,\\
{} & {} & infarction,\ inflammation \\
\hline
1 & 240-279 & cyst,\ cystic,\ nontoxic,\\ 
{} & {} & polyclonal,\ monoclonal \\
\hline
2 & 460-519 & cavity,\ spasm,\ pseudomonas,\ chest,\ lungs \\
\hline
3 & 800-999 & response,\ high,\ fracture,\ pelvis,\ superficial \\
\hline
4 & 580-629 & fistula,\ chronic,\ bilateral,\\ 
{} & {} & edema,\ loss,\ infection,\ prostate,\ cervical \\
\hline
5 & 520-579 & ml,\ wean,\ vent,\ overloaded,\ snacks,\ along \\
\hline
6 & 780-799 & shaking,\ balance,\ results,\\
{} & {} & cefipime,\ nontender,\ green,\ glucose,\ blood \\
\hline
7 & 280-289 & thrombocytopenia,\ thalassemia,\ mesenteric,\
count \\
\hline
8 & 001-139 & g,\ mycobacterial,\ leukoencephalopathy,\\\
{} & {} & neurologic,\ intrathoracic \\
\hline
9 & 320-389 & rest,\ lvot,\ tablet,\ double,\ patient,\ gradient \\
\hline
10 & 290-319 & disorders,\ stress,\ anorexia,\ mental,\ mood \\
\hline
11 & 760-779 & neonatal,\ tachypnea,\ cytomegalovirus,\ extreme,\ baby \\
\hline
12 & 680-709 & eczema,\ nail,\ rash,\ lupus,\ psoriasis \\
\hline
13 & 710-739 & multiple,\ extremities,\ post,\\
{} & {} & neck,\ acute,\ node,\ vertebral,\ patellar \\
\hline
14 & 140-239 & outpatient,\ pancreatic,\\
{} & {} & hypertension,\ stone,\ breast,\ neurologic,\ nodular \\
\hline
15 & 740-759 & cataract,\ mouth,\ patent,\ biliary,\ partial \\
\hline
16 & 630-679 & pregnancy,\ eclampsia,\ gestation,\ normal,\ trial \\
\hline
\end{tabular}
}

\caption{Sample of MSC tagged words for each of the ICD9 codes, produced by running MSC on 500 batches of MIMIC-III test data with a batch containing 256 clinical notes.}
\label{tab:words}
\end{table}

\begin{table*}[!hbt]
\centering
\begin{tabular}{|c|c|c|c||c|c|c|c|}
\hline
Index &  ICD9 Vocabulary &    MSC &  Intersection & Index & ICD9 Vocabulary & MSC  & Intersection \\
\hline
0 &   505 &  37846 &     459 &    9 &  1240 &      6 &     1 \\
\hline
1 &   393 &  21194 &     326 &   10 &   401 &   8083 &   190 \\
\hline
2 &   385 &   5147 &     174 &   11 &   402 &  16633 &   326 \\
\hline
3 &  1284 &    233 &      48 &   12 &   258 &   4055 &    94 \\
\hline
4 &   457 &   1239 &      68 &   13 &   525 &    363 &    20 \\
\hline
5 &   576 &      6 &       0 &   14 &   563 &    133 &     8 \\
\hline
6 &   654 &     54 &       9 &   15 &   575 &   2748 &   157 \\
\hline
7 &   191 &   5281 &      77 &   16 &   560 &   6106 &   303 \\
\hline
8 &   778 &  21757 &     560 &    -- &     -- &      -- &     -- \\
\hline
\end{tabular}
\caption{Word tagging statistics: Intersection of word tags produced by MSC for each index and ICD9 vocabulary -- words found in long and short descriptions of all sub-categories of an ICD9 broad diagnosis category.}
\label{tab:word-stats}
\end{table*}

\begin{figure}[!hbt]

\centering
\begin{tabulary}{\linewidth}{CC}
  \includegraphics[width=0.25\textwidth]{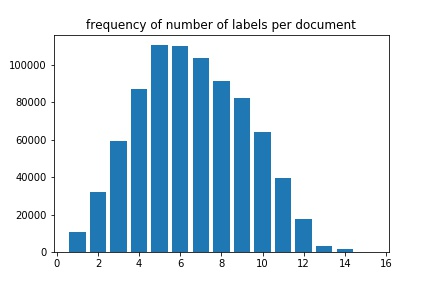}
  &
  \includegraphics[width=0.25\textwidth]{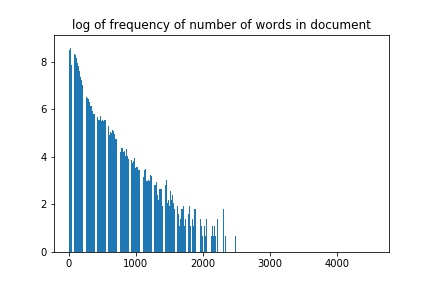}
   \\                                            
  A & 
  B \\
  \includegraphics[width=0.25\textwidth]{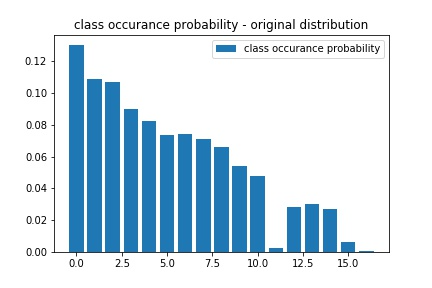}
  &
  \includegraphics[width=0.25\textwidth]{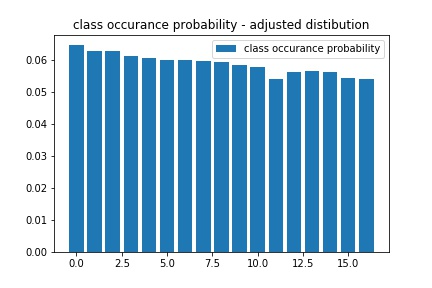}
   \\                                                
  C & 
  D \\
  \includegraphics[width=0.25\textwidth]{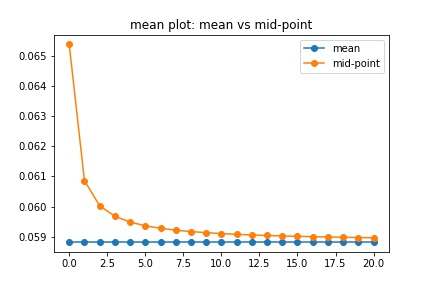}
  &
  \includegraphics[width=0.25\textwidth]{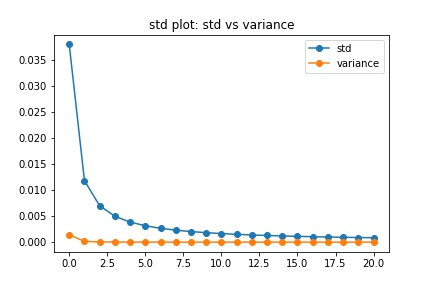}
   \\                                             
  E
  &
  F\\
 \end{tabulary}

  \caption{A: Distribution of number of fine-grained labels in the MIMIC-III data. B: Log frequency of the lengths of the clinical notes. C: Occurrence of each ICD9 broad category in the training set before balancing. D: Occurrence of each label in the training set after balancing. E: Relationship between mean (blue) and midpoint (orange) of the distribution of $A$ to the balancing parameter $c$. F: Relationship between standard deviation (blue) and variance (orange) of the distribution of $A$. Balancing occurs when the probability distribution approaches a uniform distribution at $c = 5$, and $A$ is the contingency matrix of classes and number of documents per class.}
  \label{fig:A-labels}

 \end{figure}
 
We limit the labels to diagnoses as they are relevant to making decisions about future treatments for the patient with the given history, and exclude medical procedure labels. ICD9 tags exist at four levels of detail, with the top level of the tree containing seventeen categories, as shown in Figure \ref{fig:legend}. We do not include the more generic supplementary E and V ICD9 codes which relate to causes of injury and supplemental classification because they are not the main diagnoses. The tree leaves (lowest level) contain millions of categories, at which level clinical notes are generally labeled. 
Each node at any level of the tree has a short and long description, which we only use at the evaluation stage.

We use all types of clinical notes in MIMIC-III shown in Table \ref{tab:categories} in order to incorporate various health professions and stages such as radiology, nursing and discharge summary.

\subsection{Preprocessing Notes and Labels}
We map fine grain nodes of the ICD9 tree to the coarse grain, for example 11.3 becomes 001-139. In addition, we roll up all short and long descriptions at various levels of the tree into accumulated short and long descriptions of the coarse grain nodes. As a result, each top level node has a code, title, and a list of descriptions.

We  retain all individual clinical notes of a patient covering multiple visits, and relabel all the MIMIC-III clinical note fine grain labels as coarse grain labels. We preprocess the notes by removing stop words, numbers and punctuation, and construct a vocabulary list of size $93,746$ based on all the words that occur in all the preprocessed clinical notes in the database, rather than limiting it to the most frequently occurring words, to ensure coverage. Figure \ref{fig:A-labels} part A and B shows the frequency distribution of lengths of processed clinical notes and the distribution of the number of fine-grained labels per note. 

Finally, we represent a document multi-label as a vector of zeros and ones with ones indicating membership in a class where an index in the vector represents a class. We use the pretrained word2vec \cite{mikolov2013distributed} embeddings on pubmed and PMC\footnote{\url{http://evexdb.org/pmresources/vec-space-models/}} \cite{moen2013distributional} with vocabulary of size $93,746$, and embedding size $200$. MSC trains the embeddings to customize to the clinical notes. 

\subsection{Training, Validation and Test Sets}

We use a 60, 20, 20 split to partition the MIMIC-III clinical notes to the training, validation and testing sets. In order to reduce the bias in training favoring most frequently occurring labels, we balance the training set through uneven replication of notes to have uniformly distributed label frequencies. 

We conduct this balancing as follows: let $x$ of size $M$ denote vector of replication of $M$ documents, and $r = c \times 1_N$ the desirable frequency of $N$ ICD9 labels. We solve for $x$ by minimizing $||Ax - r||^2_2$ with the constraint that $x$ is non-negative, where $A$ is the contingency matrix of the labels of each document with shape $N \times M$ where $N$ is the number of classes and $M$ is the number of documents. The multiplier $c$ is an integral multiple of the maximum number of label repetitions aggregated over the clinical notes in the training set: $c = k \times max\_rep$ where  $max\_rep = max(l)$ and  $l = \frac{\sum_M{A}}{\sum_{N,M}{A}}$ which denotes the frequency of label occurrence in the set of training data. 

The target $c$ is one in which $l$ approaches a uniform distribution. We start with a value of $c=1$ and compare the mean and variance to the mean and variance of a reference uniform distribution, $\frac{(a + b)}{2}$ and $\frac{(a - b)^2}{12}$ where $a$ is the probability that the first label occurs and $b$ is the probability that the last label occurs. Figure \ref{fig:A-labels} parts E and F plots the mean and standard deviation (std) and indicates that a value of $c = 5$ yields a good approximation to a uniform distribution. Figure \ref{fig:A-labels} parts C and D show the label occurrence distribution in the training set before and after balancing. 

\section{Approach}

Our premise is that classification of a word in a word sequence  depends on the word itself, forward words, reverse words, classification of forward words, and classification of reverse words. The dependence on context generally decreases with distance from the word itself. 
Similarly, classification of a phrase in a phrase sequence depends on the phrase itself, forward and reverse phrases, and classification of forward and reverse phrases. 
Classification of a document is dependent on classification of all phrases cumulatively without regard to location of a phrase. It is a transformation of a sequence of phrase classifications to document label. Segment coloring is determined from phrase allocation probabilities described in detail in Section \ref{sec:segment_coloring}. 

Two issues arise concerning a phrase: (1) what is it delimited by; (2) what is the mapping of word classification to phrase allocation. We solve the first problem by assuming that a typical phrase consists of five words and creating a sliding window of length five over words to specify phrase sequences. We solve the second problem by assuming that the mapping is well modeled by a GRU. 

The above description is sequential proceeding from word to phrase to document. As the labels available at training stage are at the document level, the reverse direction constraints are also implemented imposing restrictions of document labels on phrase allocation and restrictions of phrase allocation on word classification. 

\section{Model Design}

\begin{figure*}[!hbt]
  \centering
  \includegraphics[width=1\textwidth]{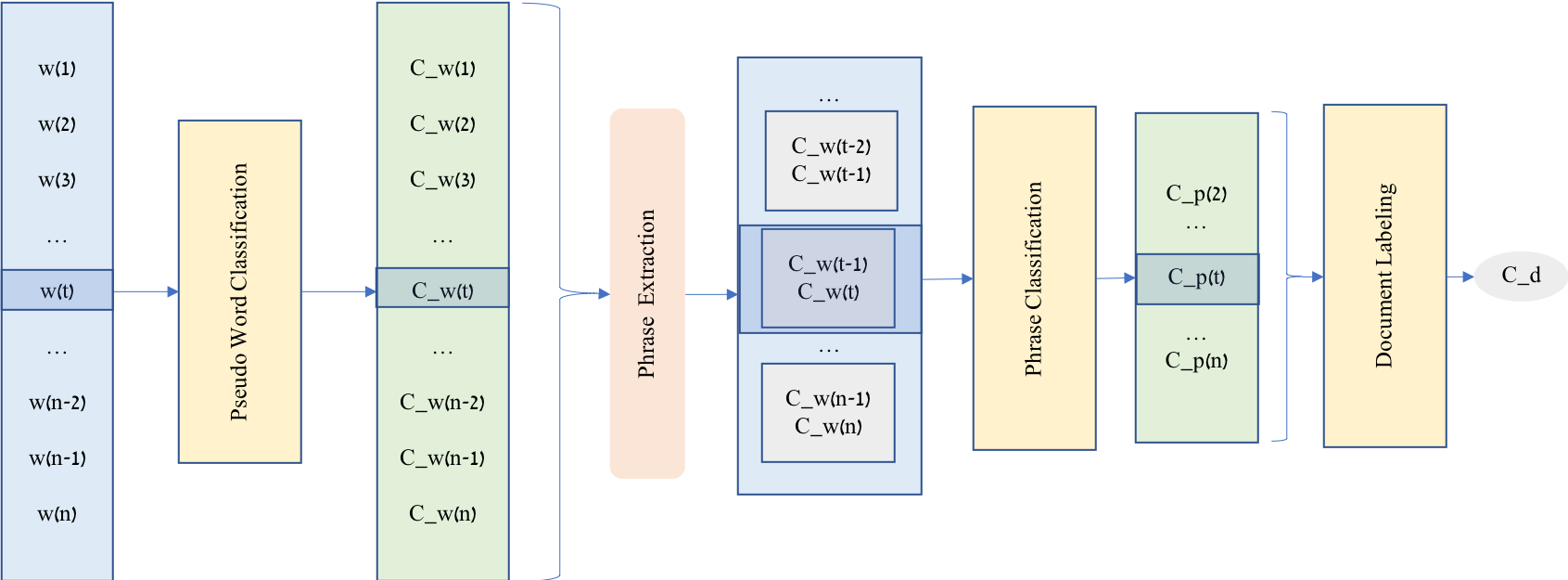}
  \caption{Model Architecture: subsystems are tied sequentially to produce word categorization, phrase allocation and document labels. Notation: t is index, w is word, n is length of a document, C\_w is the classification of a word, C\_p is the classification of a phrase, C\_d is the classification of the entire document. For illustration purposes, the size of a phrase is two.}
  \label{fig:model}
  \vspace{0.65em}
\end{figure*}

\begin{figure}[!hbt]
  \centering
  \includegraphics[scale=0.48,angle=0]{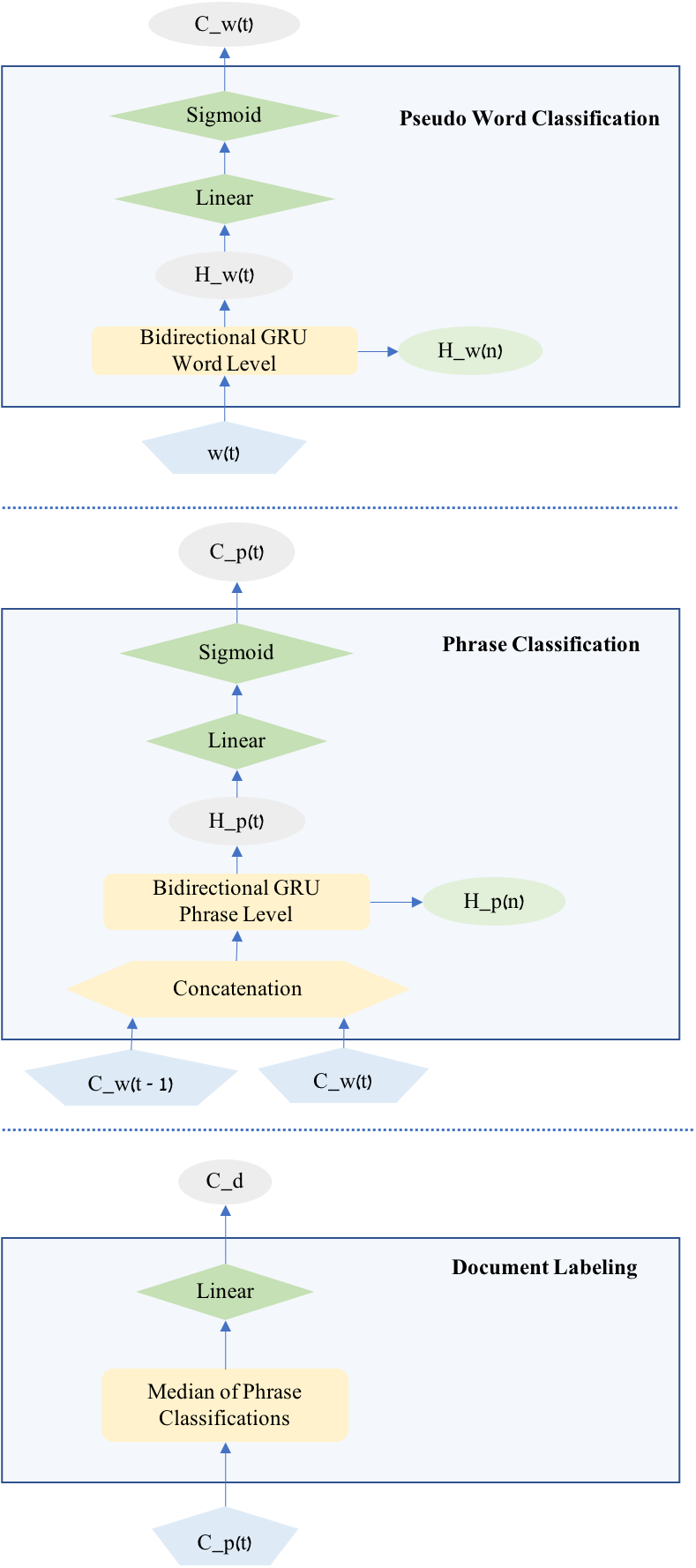}

\caption{Model architecture. TOP: pseudo classifies each word in the document; MIDDLE:  allocates each phrase in the document based on pseudo word classifications; BOTTOM: classifies the entire document based on phrase allocations. For illustration purposes, the length of a phrase is two.}
\label{fig:model_sub_models}
\end{figure}

Our design consists of architectures at serially linked word, phrase, and document levels that imposes bidirectional constraints on one another, and pipelines input document into word classifications, phrase allocations and document labels. Figure \ref{fig:model} schematizes the architecture and Figure \ref{fig:model_sub_models} shows how each sub-component computes its output. 

The word module maps between sequence of words in a document and sequence of their multi-labels, phrase module maps between multivariate sequence of delayed word multi-labels and window-shifted phrase multi-labels. The phrase multi-labels are mapped to document level classification using the median of the phrase allocation.

In Figure \ref{fig:model_sub_models} (top), we treat a word as a unit and because of the bidirectionality of the GRU, the classification probability of a word at time $t$ is directly dependent on the word at time $t$ and the classification of the preceding and succeeding words. Similarly, in \ref{fig:model_sub_models} (middle) the phrase unit is obtained by a sliding window of size five on the word classifications. And the phrase classification at time $t$ is directly based on the phrase at time $t$ and classifications of preceding and succeeding phrases. Consequently, a word/phrase classification is indirectly dependent on classification of previous and future classifications of words/phrases, with this dependence becoming weaker the further away from the unit -- as a correct reflection of text dependencies.

\subsection{Word Architecture}

The word architecture takes serially as input the sequence of words in a document, $w(t)$, and outputs pseudo classification of a word into the broad ICD-9 codes, $C_w(t)$. We call the classification pseudo because it is only valid in the context of the word and is not very meaningful as context independent classification.

A word is translated into its embedding, and then the sequence is processed by bidirectional GRU (gated recurrent neural network) which classifies a word based on itself and its forward and backwards context. Each item in the output sequence corresponding to a word is then inputted to a linear layer followed by a sigmoid to compute the probabilities of a word belonging to each of the various classes. 

\subsection{Phrase Architecture}

The phrase architecture takes serially as input a sequence of word classifications $C_w(t)$ and outputs serially the sequence of phrase classifications, $C_p(t)$. It first unfolds the input into the multivariate sequence $C_w(t),C_w(t+1),C_w(t+2),C_w(t+3),C_w(t+4)$, for a phrase of size five. We choose five as the typical length of an English phrase. 

A bidirectional GRU then transforms the multivariate sequence to a univariate sequence which is then inputted to a linear layer followed by a sigmoid to get the probabilities of a phrase belonging to each of the various classes. Incorporation of all words in a phrase of length five is enabled by the unfolding action, thereby creating a correspondence between phrase classification and pseudo classification of its word constituents.

\subsection{Document Architecture}

The document architecture takes the median of the phrase classification probability vectors, thus getting the approximation of probabilities of the document belonging to each of the classes. We choose median over mean and mode as it is a robust measure of centrality. 

\subsection{Segment Coloring}
\label{sec:segment_coloring}

A segment is a contiguous string of words with the same tag. A tag is a mildly context-free single-label classification of the word based on the phrase allocation probabilities of the phrases encompassing the word.

Segment m is defined as $S_m\ =\ [w_j,\ ... ,\ w_k]$, with $R(j)\ =\ R(j+1)\ =\ ...\ =\ R(k)$, where  $R(t)$ is a tag of word $w_t$ for $t\ =\ 1:n$; and $n$ is the size of the clinical note. At the delimits $[j,k],\ R(j-1)\ !=\ R(j)$ and $R(k)\ !=\ R(k+1)$.

Tag $R(t)$ is computed as follows: 
\begin{enumerate}
    \item Identify set of phrases covering $w_t$ : ${p(t),\ ... ,\ p(t+4)}$ where phrase $p(t)$ has five words. 
    \item Choose the label with highest probability for each phrase: $[argmax(C_p(t)),\ ... ,\ argmax(C_p(t+4))]$, where $C_p(t)$ is the computed sigmoid multi-label probability of phrase $t$. We ignore labels that have probability $< 0.5$.
    \item Determine tag of word $w_t$: 
$R(t)\ =\ argmax(C_p(t))$ if $argmax(C_p(t))\ == argmax(C_p(t+1))\ == \ ... ==\ argmax(C_p(t+4))$ else $R(t) = 'no\ tag'$
\end{enumerate}

The segment is colored according to the tag color shown in Figure \ref{fig:legend}. Listing of words belonging to each ICD9 code is compiled based on the word tags. Table \ref{tab:words} shows a sample of tagged words for each of the 17 ICD9 codes and Table \ref{tab:word-stats} shows the evaluation measured as intersection between the ICD9 vocabulary and the word tags.

% ------------------------

\section{Supervised Learning}

We use AMSgrad \cite{reddi2019convergence} to optimize the model parameters by minimizing the  binary cross entropy loss (BCELoss) which is a loss typically used when measuring the error of reconstruction. 
The parameter values used in training are: 256 batch size, 300 training batches, $\sim$ 1000 epochs with a 0.001 learning rate, other parameters are the default used. MSC uses a hidden layer of size 20 for word GRU and 10 for phrase GRU. 

The loss per batch is computed by the following equation: 
$L\_b(x, y)$ =  average([$l\_n$ for $n$ in range($1$, $N$)]) where $N$ is the number of elements in a batch $l_n = - w_n \times [y_n \times log(x_n) + (1-y_n) \times log(1-x_n)]$ where $l_n$ is the loss for a single output in the batch. 

\section{Performance}

\begin{table}[!hbt]
\centering

\begin{tabular}{|c|c|c|c|c|c|c|c|c|}
\hline
Index & P1 & P2 & P3 & P4 & P5 & P6 & Avg \\
\hline
0 & 1 & 1 & 1 & 1 & 1 & 1 &  100\% \\
\hline
1 & 1 & 1 & 1 & 1 & 1 & 1 &  100\% \\
\hline
2 & 1 & 1 & 1 & 1 & 0 & 0 & 66.7\% \\
\hline
3 & 0 & 1 & 0 & 0 & 0 & 0 & 16.7\% \\
\hline
4 & 0 & 1 & 0 & 0 & 0 & 0 & 16.7\% \\
\hline
5 & 0 & 1 & 0 & 0 & 0 & 0 & 16.7\% \\
\hline
6 & 0 & 1 & 0 & 0 & 0 & 0 & 16.7\% \\
\hline
7 & 1 & 1 & 1 & 1 & 1 & 1 &  100\% \\
\hline
8 & 1 & 1 & 1 & 0 & 1 & 1 & 83.3\% \\
\hline
9 & 0 & 1 & 0 & 0 & 0 & 0 & 16.7\% \\
\hline
10 & 1 & 1 & 1 & 1 & 1 & 1 & 100\% \\
\hline
11 & 1 & 1 & 1 & 1 & 0 & 1 & 83.3\% \\
\hline
12 & 1 & 1 & 1 & 0 & 1 & 1 & 83.3\% \\
\hline
13 & 1 & 1 & 0 & 1 & 1 & 1 & 83.3\% \\
\hline
14 & 1 & 1 & 0 & 0 & 0 & 0 & 33.3\% \\
\hline
15 & 0 & 1 & 0 & 1 & 1 & 1 & 66.7\% \\
\hline
16 & 1 & 1 & 1 & 1 & 1 & 1 & 100\% \\
\hline
mn &  -- & -- & -- &  -- & -- & -- & 63.7\% \\
\hline
md & -- & -- & -- & -- & -- & -- & 83.3\% \\
\hline
\end{tabular}

\caption{Assessment of MSC coloring output by medical practitioners P1 - P6. Match between their category assignment and the legend is indicated by 1. The mean (mn) and median (md) scores are  $63.7\%$  and $83.3\%$ respectively.}
\label{tab:qual}
\end{table}{}

To evaluate performance, the trained MSC model is run on a test set of 500 batches with a batch size of 256 clinical notes from MIMIC-III. We compute performance of MSC's coloring capability through scoring of coloring results by medical professionals as described in Section \ref{sec:eval}. 

In addition, we examine the word tags as detailed in Section \ref{sec:word_tags_evaluation} and compute the F1-score of document multi-labeling. MSC's document multi-labeling achieves scores of: $64\%$ micro-f1, $67\%$ micro-precision, and $61\%$ micro-recall on the test set. 
\subsection{Word Tagging}
\label{sec:word_tags_evaluation}

We examine the word tags produced by MSC on the entire test set by comparing them to ICD9 tag words. The reference is created by deriving sets of words for each label from the ICD9 code descriptions -- long and short for all sub-categories related to the broader category. 

We use set intersection between our word tags of the test set and the reference as a metric for evaluation of word tagging and as a proxy for segment coloring. Table\ref{tab:word-stats} lists the intersection statistics and Table \ref{tab:words} shows a sample words related to the ICD-9 categories as tagged by our model. Note  that some of the indices have a reduced number of common words at less than 50: 3 (injury and poisoning), 5 (digestive system), 6 (symptoms, signs, and ill-defined conditions), 9 (nervous system and sense organs), 13 (musculoskeletal system and connective tissue), and 14 (neoplasms).

\subsection{Segment Coloring Scoring}
\label{sec:eval}

Figure \ref{fig:example_colored} presents three clinical notes from the test set after being colored by MSC and Figure \ref{fig:legend} shows the coloring scheme used when coloring the segments.

% We would like a scoring method that is objective while providing flexibility to medical practitioners in evaluating coloring results.
An unstructured evaluation of performance would entail asking medical practitioners to color segments of the prepossessed clinical notes according to the ICD9 category coloring scheme; and then evaluating the discrepancies against MSC's output. However, this poses difficulties in penalizing presence or absence of segments and assigning partial scores to their delimitations which are subjective. To overcome these difficulties and objectively evaluate coloring performance, we presented 40 MSC colored clinical notes along with 50 tagged words for each color to six medical practitioners without the legend. 

We ask the medical practitioners to assign one of the 18 descriptions (17 ICD9 and one generic) to a color based on the colored clinical notes and tagged words. No legend or descriptions of colors are provided to the evaluators. Table \ref{tab:qual} shows the detailed evaluation. The mean and median of binary match between MSC's color and medical practitioners color is $63.7\%$ and $83.3\%$ respectively. The mean score is $81.8\%$ for the 11 indices that have word tag with more than 50 words overlapped. 

% Note the correlation between word tagging and  matching of descriptions by medical professionals to the legend

\section{Conclusion}

We solve the problem of identifying, delimiting and coloring medical segments of a clinical note with the color of a segment corresponding to one of the 17 top-level diagnosis ICD9 categories. Coloring provides a medical specialist focused segments tailored to his/her specialization while providing context, thereby insuring that important diagnoses are not overlooked in a long medical note. In comparison, the attention problem highlights individual words that effect labeling, and the labeling problem outputs the multiple labels.  

% and as byproducts tags words and multi-labels clinical notes with ICD-9 diagnosis categories

Segment coloring performance is computed by scoring the categories assigned to the colors against the association of colors to categories made by medical professionals. The mean score obtained is $63.7\%$ and the median score is $83.3\%$. The multi-labeling byproduct of our coloring model -- MSC -- achieves a micro score of $64\%$ F1-score, $67\%$ precision, and $61\%$ recall for documents; which compares favorably to recently published values of $52.4\%$ F1-score for the attention problem model -- CAML -- which uses attention to highlight the words affecting the labeling. 

The design of MSC models at the three levels of words, phrases, and document to incorporate information at various levels and their dependencies based on natural language structure. We utilize bidirectional GRUs to model word dependencies and its relationship to phrases and consequently document.

Future work consists of fine tuning MSC to reduce the proportion of generic segments allocated to the categories. One way to achieve this is by using word lists for each ICD-9 category based on domain knowledge for pre-training the embeddings.

% \section*{Acknowledgments}

\newblock

\bibliographystyle{acl_natbib}
% \bibliography{template}  %%% Uncomment this line and comment out the ``thebibliography'' section below to use the external .bib file (using bibtex) .

% Uncomment this section and comment out the \bibliography{references} line above to use inline references.

\end{document}